\pdfoutput=1

\documentclass[11pt]{article}

\usepackage[preprint]{acl}
\usepackage{float}

\usepackage{times}
\usepackage{latexsym}

\usepackage[T1]{fontenc}

\usepackage[utf8]{inputenc}

\usepackage{microtype}
\usepackage{subcaption}

\usepackage{inconsolata}

\usepackage{graphicx}

\usepackage{amsmath}

\title{\textsc{RAGViz}: Diagnose and Visualize Retrieval-Augmented Generation}

\author{Tevin Wang\text, Jingyuan He, Chenyan Xiong\\
 School of Computer Science, Carnegie Mellon University \\
 Pittsburgh, PA 15213 \\
  \texttt{tevin@cmu.edu, jingyuah@cs.cmu.edu, cx@cs.cmu.edu} \\}

\begin{document}
\maketitle

\begin{abstract}
Retrieval-augmented generation (RAG) combines knowledge from domain-specific sources into large language models to ground answer generation. Current RAG systems lack customizable visibility on the context documents and the model's attentiveness towards such documents. We propose RAGViz, a RAG diagnosis tool that visualizes the attentiveness of the generated tokens in retrieved documents. With a built-in user interface, retrieval index, and Large Language Model (LLM) backbone, RAGViz provides two main functionalities: (1) token and document-level attention visualization, and (2) generation comparison upon context document addition and removal. 
As an open-source toolkit, RAGViz can be easily hosted with a custom embedding model and HuggingFace-supported LLM backbone.
Using a hybrid ANN (Approximate Nearest Neighbor) index, memory-efficient LLM inference tool, and custom context snippet method, RAGViz operates efficiently with a median query time of about 5 seconds on a moderate GPU node.\footnote{Our code is available at \url{https://github.com/cxcscmu/RAGViz}.
A demo video of RAGViz can be found at \url{https://youtu.be/cTAbuTu6ur4}. 
} 

\end{abstract}
\begin{figure*}
\centering
    \begin{subfigure}[t|t]{\textwidth}
        \includegraphics[width=0.5\columnwidth]{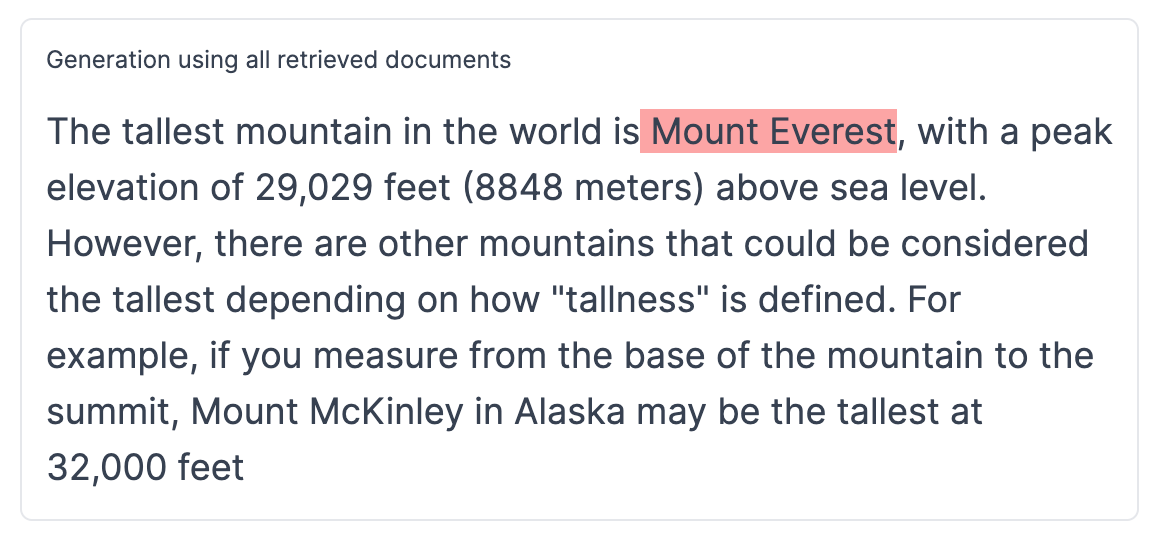}
        \includegraphics[width=0.5\columnwidth]{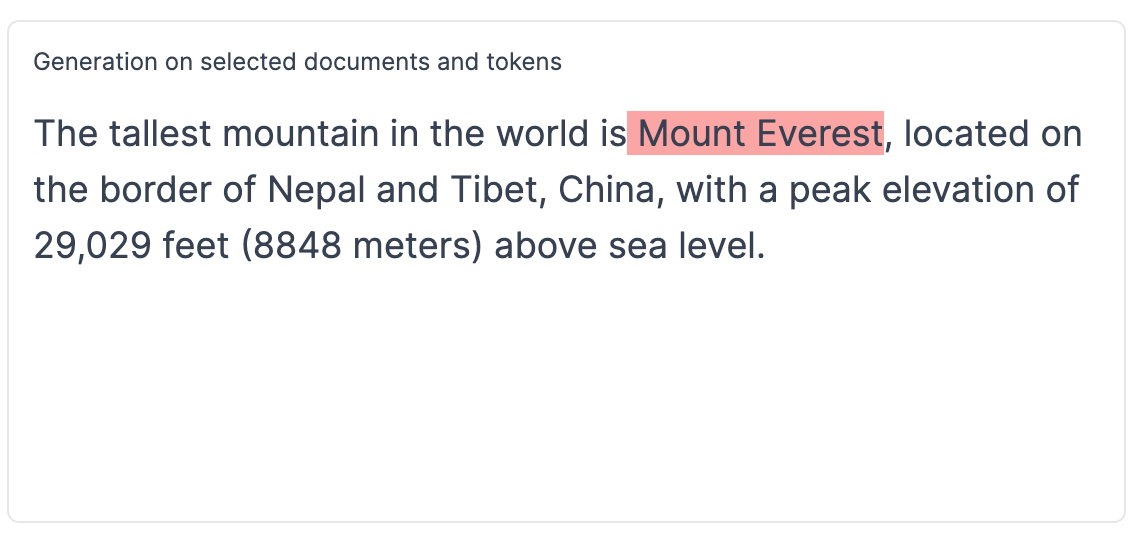}
                \caption{LLM generation for the query \textit{What is the tallest mountain in the world?} The highlighted text shows the selected tokens for attention visualization. The first generation uses both documents and the second generation uses only the second document.}
        \label{fig: python - generations}
    \end{subfigure}
    \begin{subfigure}[t]{1.0\columnwidth}
        \includegraphics[width=\columnwidth]{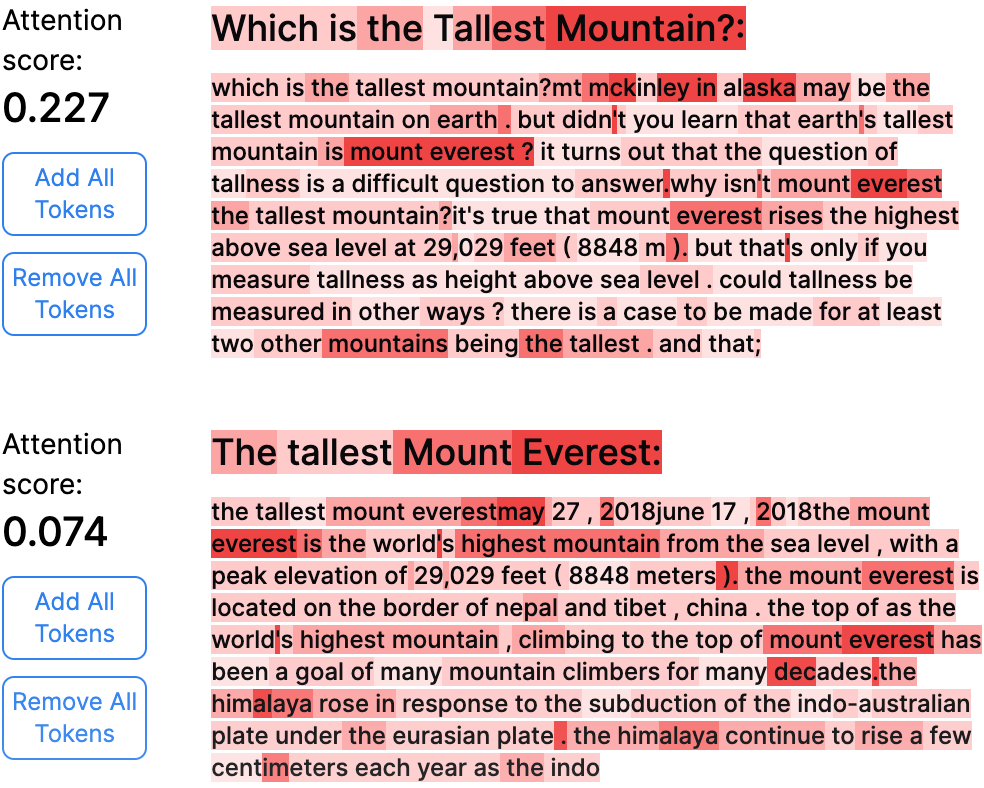}
        \caption{Initial attention visualization with both context documents.}
        \label{fig: python - scores}
    \end{subfigure}
    ~
    \begin{subfigure}[t]{\columnwidth}
        \includegraphics[width=\textwidth]{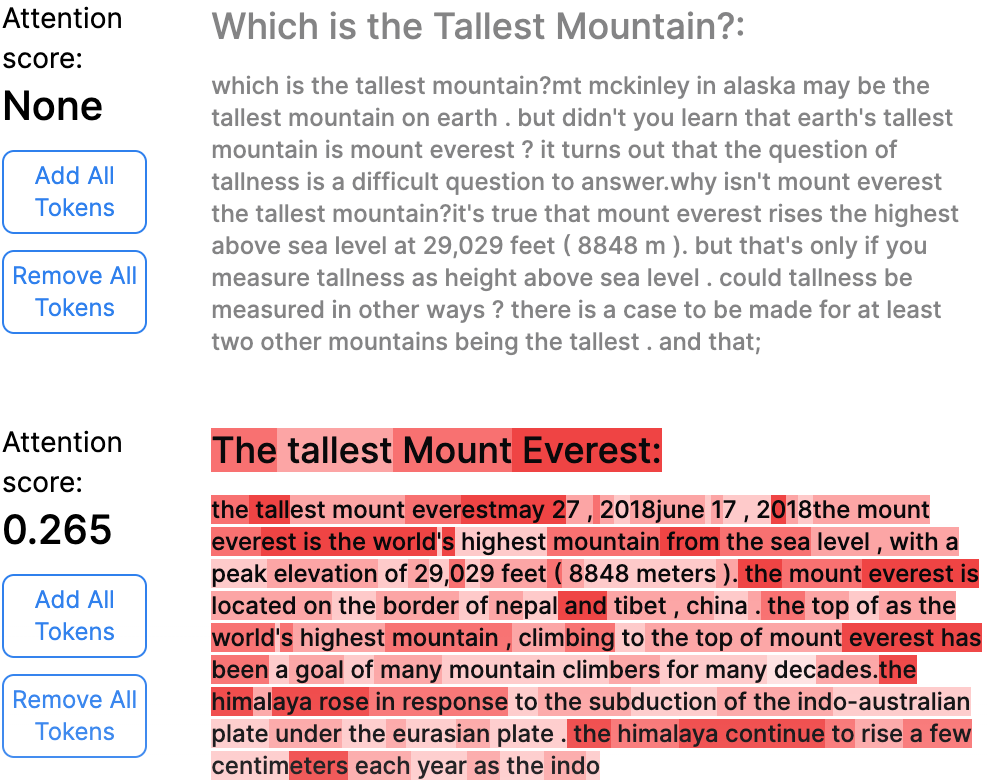}
        \caption{Attention visualization after removal of the first document.}
    \label{fig: python - after removal}
    \end{subfigure}
\caption{\textit{Attention visualization} on the \textit{selected token sequence} when using the \textit{document toggling} feature.}
\label{fig: python}
\end{figure*}
\section{Introduction}
Large language models (LLMs), such as GPT-4 \citep{openai2024gpt4technicalreport}, have revolutionized the field of artificial intelligence with their impressive language understanding and generation capabilities developed through extensive pretraining on large-scale textual data. 

A key limitation of using pretrained LLMs for zero-shot answer generation is their lack of access to domain-specific knowledge, as these models rely solely on parametric memory. The fixed knowledge derived from parametric memory often leads to hallucinations. To address this issue, \citet{NEURIPS2020_6b493230} introduces retrieval-augmented generation (RAG), a technique that leverages retrieval mechanisms to incorporate non-parametric memory, typically derived from documents retrieved from domain-specific data stores.

Various systems have been developed to deliver RAG services. For instance, \textit{OpenAI Assistants} \citep{openaiassistants} and \textit{Pinecone Assistant} \citep{pineconeassistant} are "chat-with-your-files" products that use retrieved documents as context for a chatbot. While these RAG systems offer state-of-the-art performance in grounded answer generation, they lack explainability regarding the efficacy of the context documents they use to produce those answers.

Some existing tools have been developed to improve language model explainability, such as \textit{BertViz} \citep{vig2019visualizingattentiontransformerbasedlanguage}, an open-source Python tool that provides attention visualizations for transformer models. Although such tools effectively analyze input token importance, they lack a customizable approach for analyzing the interaction between retrieved context documents and language generation.

In this paper, we propose RAGViz, a diagnostic tool designed to analyze LLM attention mechanisms on the retrieved documents that provide context to ground LLM answer generation. RAGViz’s novelty lies in its focus on the interaction between the retrieval pipeline and the language model.
RAGViz offers attention visualizations based on different levels of scoring: both cumulative attention scores on documents and individual token attention scores selected by the user. Along with document toggling, RAGViz enables users to qualitatively assess the effectiveness of retrieved documents and determine whether they contribute to hallucinations.

RAGViz's system primarily relies on CPU nodes, with the exception of a GPU node that hosts the LLM inference server. The system entry point is a web node that hosts the frontend as static content and routes queries to the main CPU node. This node forwards the query to worker nodes for document retrieval, builds the context, and sends the request to the GPU node for LLM inference. The generated answer and associated attention scores are then returned as an HTTP response to the frontend. 

RAGViz achieves efficiency through its distributed architecture and optimized LLM inference, partitioning large datasets across multiple nodes for parallel processing and faster retrieval. It uses fast inference libraries for low-latency LLM output generation. Additionally, RAGViz is customizable, allowing integration with any retrieval pipeline or attention-based language model architecture supported by HuggingFace \citep{wolf-etal-2020-transformers}, offering flexibility for diverse research needs.
\section{RAGViz Features and Use Cases}
This section first examines the innovative features of RAGViz and outlines its key benefits. Then, a few potential use cases are explored to demonstrate how RAGViz can be valuable to researchers and domain experts.

\subsection{Features}

RAGViz's system includes a few key features. One is the \textit{attention visualization on retrieved documents}. RAGViz uses token highlighting to visualize the attentiveness of any generated token sequence to input tokens, as shown in Figure \ref{fig: python - scores}. The level of attentiveness is measured by the attention score across all layers of the LLM and visualized by color magnitude. A cumulative document-level attention score is displayed to showcase the attentiveness of the generation output to each retrieved passage.

RAGViz also offers a \textit{drag-to-select user interface}. By simply dragging and selecting, users can easily inspect the cumulative attention of any token sequence, as demonstrated in Figure \ref{fig: python - generations}.

In addition to attention visualization, RAGViz provides \textit{document toggling functionality}. By toggling, users can select tokens and documents to omit when constructing the answer generation context. The newly generated answer will be shown side-by-side with the original answer to provide a comparative analysis of how adding or removing tokens and documents affects the LLM output. An example of the attention visualization changes after removing a document is in Figures \ref{fig: python - scores} and \ref{fig: python - after removal}.

Furthermore, RAGViz offers the ability to select a \textit{custom number of context documents}. Users can enter the number of relevant document snippets to retrieve from the dataset. RAGViz also includes \textit{API key authentication}, as it implements middleware functions on top of HTTP requests to ensure that requests are properly authenticated.

\subsection{Benefits}

\begin{figure*}
\centering

    \begin{subfigure}[t]{0.49\textwidth}
        \includegraphics[width=\textwidth]{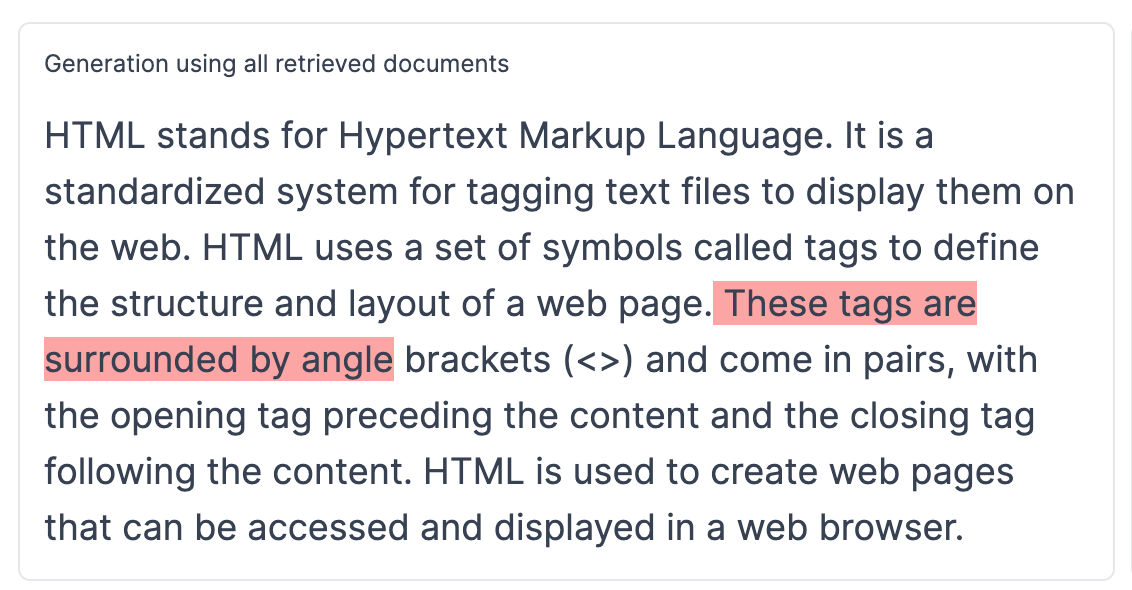}
        \caption{Initial generation with respect to the query \textit{What is HTML?} that includes unnecessary HTML tag information.}
    \label{fig: RAGViz demo - seq selection}
    \end{subfigure}
    \hfill
    \begin{subfigure}[t]{0.49\textwidth}
        \includegraphics[width=\textwidth]{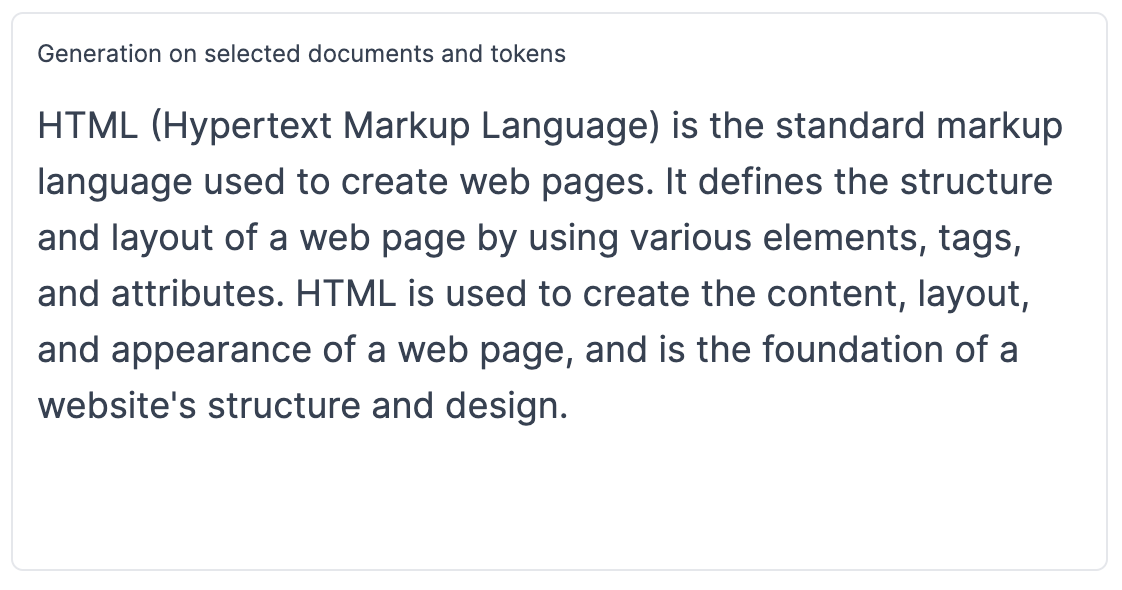}
        \caption{Response generated after the document snippet below is removed. The response is more focused and concise.}
        \label{fig: RAGViz demo - response after removal}
    \end{subfigure}

\begin{subfigure}[t]{\textwidth}
\includegraphics[width=\textwidth]{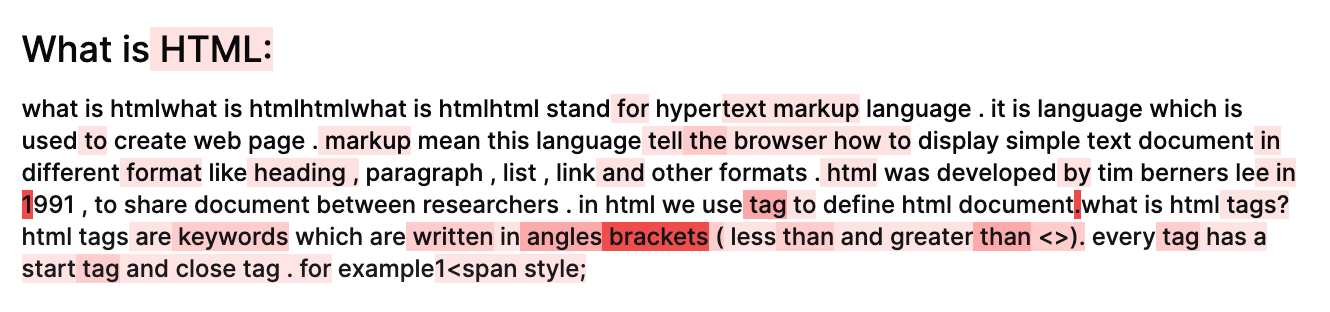}
\caption{A document with HTML tag information that the LLM is attending on to generate the first response. }
\end{subfigure}

\caption{ A demo of RAGViz showcasing RAGViz's ability to identify and debug external hallucinations. }
\label{fig: RAGViz demo}
\end{figure*}

Through the features described, RAGViz provides several key advantages.

Firstly, RAGViz enables precise \textit{document efficacy diagnosis} through attention-based visualizations. By examining how LLMs allocate attention across different retrieved context documents during generation, users can assess the quality and relevance of the retrieval process. This helps identify which document contributes meaningfully to the generated output and which may lead to irrelevant or hallucinated information.

Secondly, the system's \textit{multi-level attention visualizations} offers flexibility for users to inspect attentiveness at various levels of granularity. With its intuitive drag-to-select interface, users can analyze attention not only at the token level but also at the phrase or sentence level. This allows for a deeper exploration of how specific sections of the text influence the model’s output.

Another significant advantage of RAGViz is its ability to support \textit{iterative experimentation} with document context. Through its document toggling functionality, users can modify the input context by adding or removing specific documents, and then compare the resulting generation side-by-side. This iterative approach helps in understanding how changes to the context impact the final output, using attention scores as a heuristic for evaluation.

In addition, RAGViz simplifies \textit{comparative analysis} by displaying original and modified outputs alongside their corresponding attention scores. This side-by-side visualization allows users to observe how variations in input documents affect the generation, providing valuable insights into the interaction between retrieval and generation.

RAGViz enhances \textit{retrieval precision testing} by allowing users to adjust the number of documents retrieved for a query. This feature enables diagnostic testing to determine whether fewer or more documents are necessary for the model to generate accurate and well-grounded responses. 

RAGViz is also \textit{private and secure}. Its basic API key authentication functionality restricts access and ensures that datasets and models are protected.

\subsection{Example Use Cases}

RAGViz presents several use cases for researchers and developers working with RAG pipelines. We highlight a few of these use cases.

One use case is to analyze the \textit{interpretability of attention mechanisms} within large language models. A key need in RAG systems is to understand how context is leveraged to produce grounded results. RAGViz provides a novel tool that enables researchers to explore the distribution of attention across different parts of the retrieved snippets, offering insight into how context documents influence the generation process.

Another application is to \textit{design and evaluate new retrieval mechanisms} tailored to RAG. The ability to visualize attention on documents in RAGViz provides researchers with a powerful method to iterate and refine the retrieval process, facilitating the development of more effective retrieval strategies to better support LLMs in grounding their outputs.

RAGViz serves as a valuable tool for \textit{debugging RAG pipelines}, particularly in diagnosing the sources of hallucinations. RAGViz can help differentiate between hallucinations caused by the retrieved documents or those stemming from the LLM’s internal parameters. For instance, if a hallucination occurs when the model shows a high concentration of attention on specific context documents, it is likely that the source of the error lies within the retrieved data. Conversely, if the attention is not focused on any particular document, the issue may originate from the model's own parametric memory.

Additionally, RAGViz enables \textit{domain experts} to assess the \textit{effectiveness of various data stores} for RAG-based systems. By visualizing the attention levels on documents retrieved from different data stores, users can evaluate which data stores are most suitable for addressing domain-specific queries, offering critical insights into the alignment between the data store and the model’s generation.

\subsection{Examples}
In this section, we showcase how RAGViz can help debug RAG pipelines by identifying hallucinations from parametric and non-parametric memory.
\begin{figure}[H]
\centering
\includegraphics[width=\columnwidth]{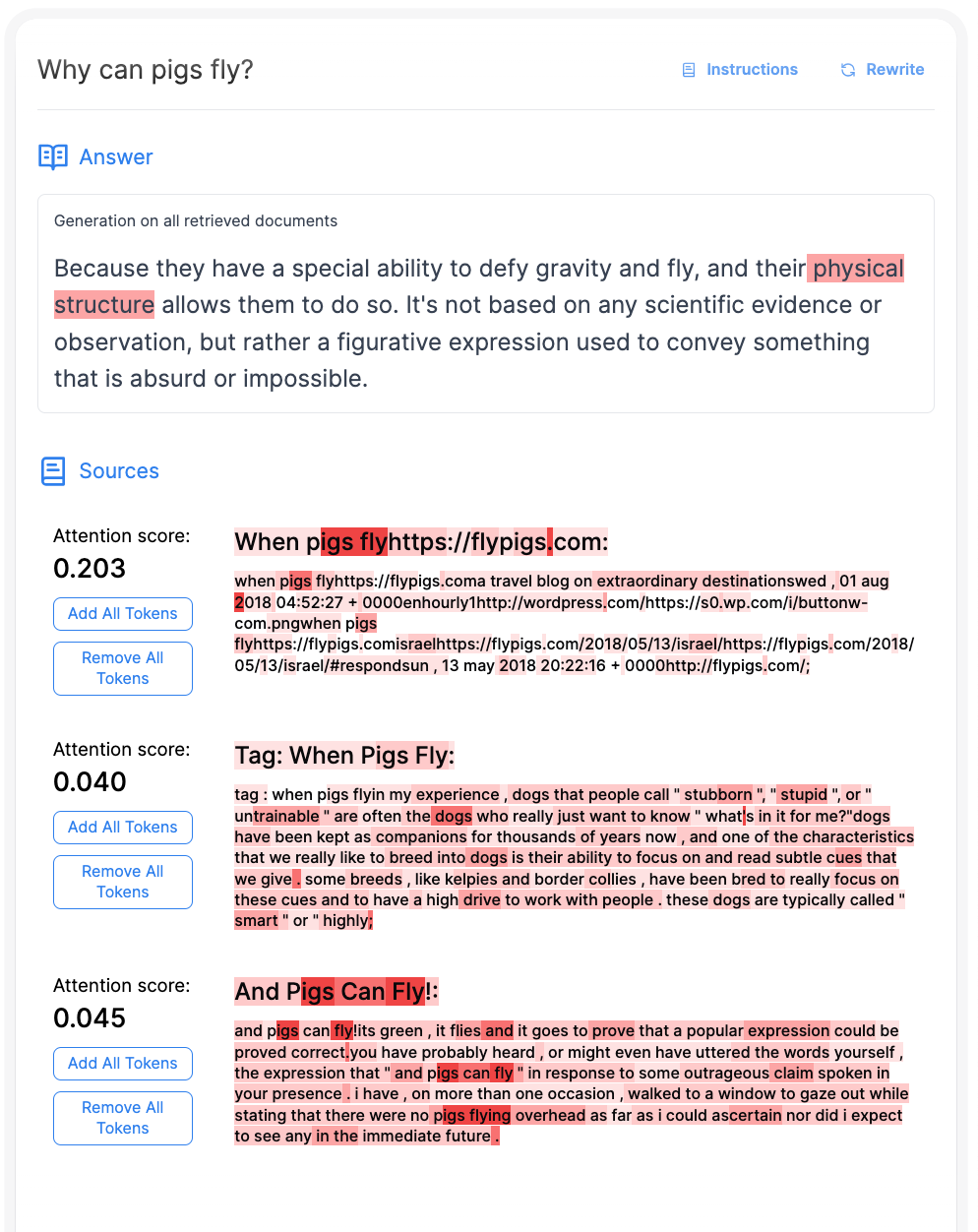}
\caption{ Visualization for query \textit{Why do pigs fly?}. The highlighted generation is not grounded by any context documents, demonstrating internal hallucination. 
}
\label{fig: RAGViz demo 2}
\end{figure}
Consider the query \textit{What is HTML?}. The generated outputs and RAGViz visualizations for such query are shown in Figure \ref{fig: RAGViz demo}. Users might utilize this query to gain an understanding of HTML and can use RAGViz to identify the context document providing the LLM with unwanted information, such as the HTML tag syntax. Figure \ref{fig: RAGViz demo} shows that the tag syntax in the generation is being influenced by a document that mentions the HTML tag, indicating that the hallucination is caused by external (non-parametric) memory. After removing this document and regenerating, the new output becomes substantially more focused on describing the concept of HTML rather than the specifics of syntax.

Figure \ref{fig: RAGViz demo 2} displays an example of internal hallucination. RAGViz's attention visualization reveals that the generated phrase "physical structure" is not grounded by any retrieved documents but stems from the LLM's internal (parametric) memory. In this way, RAGViz provides qualitative insights into why different parts of the output were generated.


\section{System Architecture}
This section introduces RAGViz's system architecture and its query pipeline. The system has four main components: the ANN (Approximate Nearest Neighbor) index for dense retrieval, the backend server, the LLM inference server, and the frontend user interface. These components are implemented separately to allow for configurability. RAGViz's system is originally designed for use with a job scheduler like SLURM \citep{10.1007/10968987_3}.
\begin{figure*}[t]
\centering
\includegraphics[width=\textwidth]{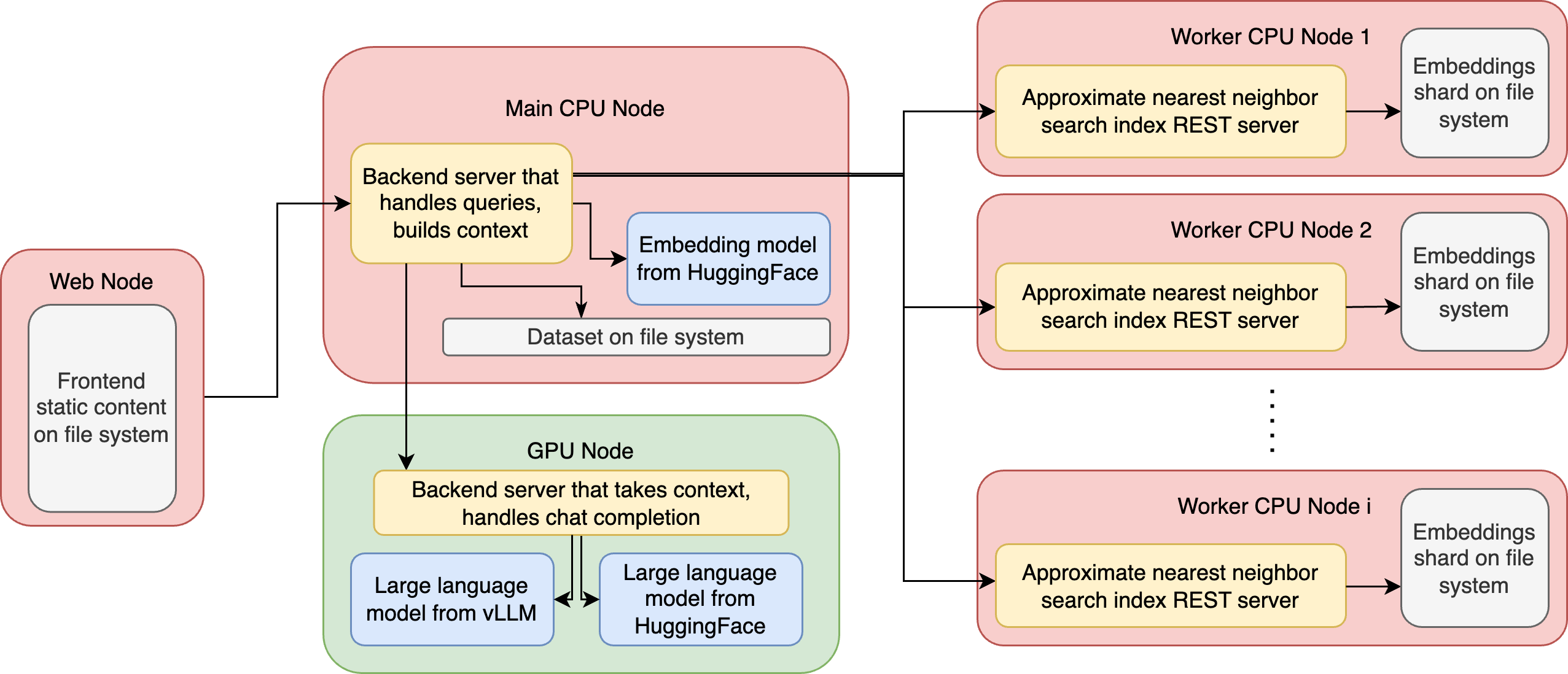}
\caption{ High-level view of RAGViz's system architecture. The arrows within nodes represent the model use or filesystem reads. The arrows between nodes represent REST API calls. Queries are routed to each of the approximate nearest neighbor search REST servers and then reranked by the context building backend server.
}
\label{fig: architecture}
\end{figure*}

\subsection{Dense Retrieval} In dense retrieval, queries and documents are encoded into high-dimensional feature vectors, also known as embeddings. A similarity search using metrics like cosine similarity or inner product is then performed to determine the nearest neighbors of a particular query vector. Significant research efforts have focused on various Approximate Nearest Neighbor Search (ANNS) indexing algorithms \citep{NIPS2004_1102a326}, which reduce search time by approximating the exact K-Nearest Neighbor search (KNNS).

For large-scale datasets, storing the embeddings and hosting an index for ANNS is often unfeasible on a single machine. RAGViz solves this by using a distributed system, where partitions of the set of embeddings are individually indexed and stored on the SSDs of separate nodes, represented in Figure \ref{fig: architecture} as worker CPU nodes 1 through $i$. The worker nodes each hosts a REST API that accepts query embeddings and returns the approximated top-$k$ nearest neighbors in the form of dataset indices.

\subsection{Context Builder} These REST API servers receive requests from the context-building backend server, which handles all the logic for constructing the language model context. Its responsibilities include loading the embedding model, managing backend logic, and storing the full corpus. This context builder is represented in Figure \ref{fig: architecture} as the main CPU node. Once queries are received and processed by authentication middleware, they are encoded into embeddings and routed to all worker nodes to perform ANNS. The top documents retrieved from the index at each worker CPU node are then reranked to return the final top $k$ nearest neighbors of the query in the whole dataset. 

\begin{figure}[t]
\centering
\includegraphics[width=\columnwidth]{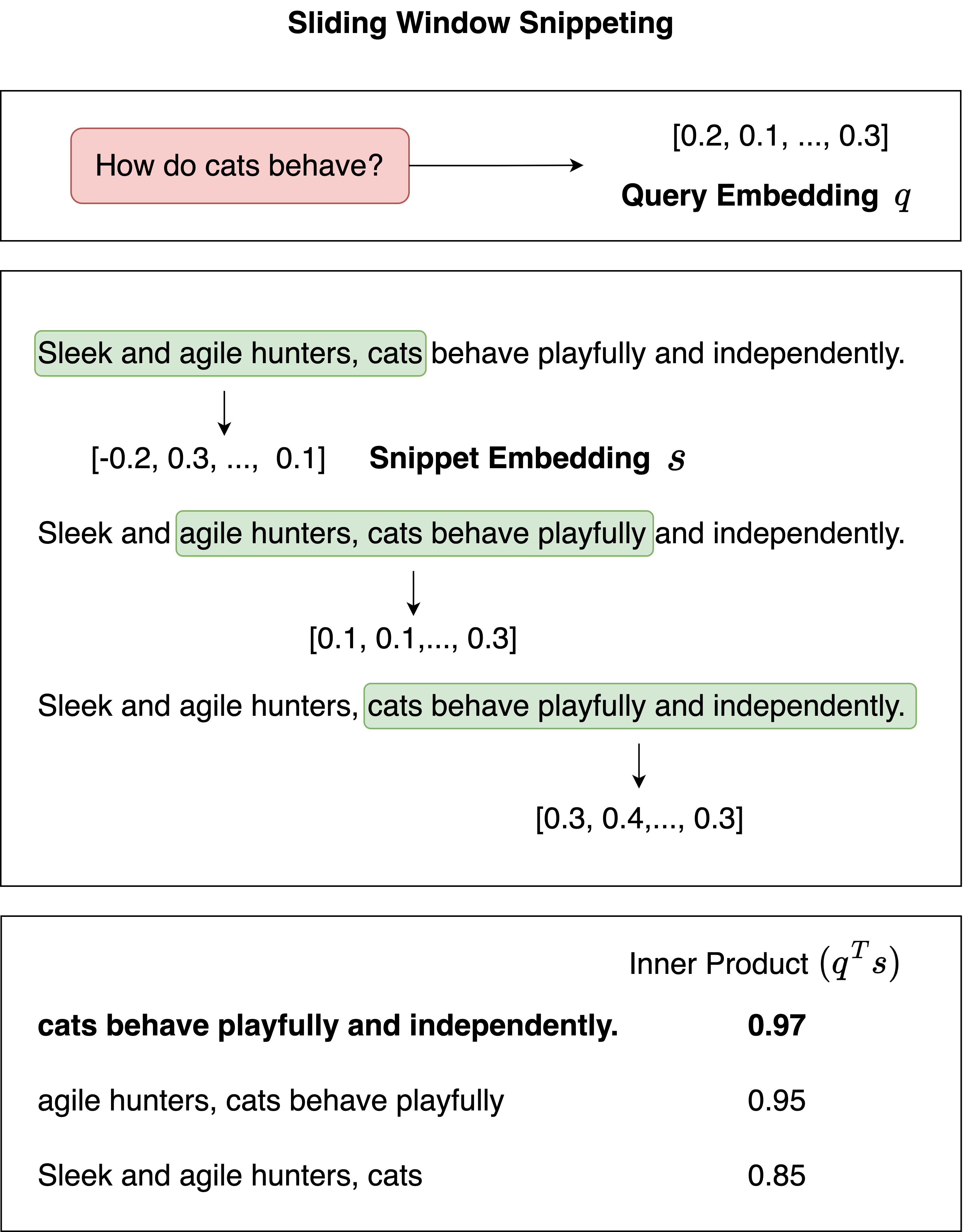}
\caption{\label{fig:snippeting}A demonstration of sliding window snippeting with a window size of $5$ and a stride of $2$. The sliding window method chooses the snippet with the highest inner product similarity. Conversely, the naive first method always selects the first window shown in green.}
\end{figure}

Once these documents are retrieved, a snippeting technique is applied to extract the portion of the document relevant to the query. RAGViz provides two document snippeting methods: naive first and sliding window. The naive first method represents a document by its first 128 tokens. The sliding window method embeds windows of 128 tokens from the document into vectors and uses the window whose encoded vector has the highest similarity with the query to represent the corresponding document. Figure \ref{fig:snippeting} shows a diagram of the sliding window method. This method increases latency in exchange for better document representation, based on the assumption that embedding similarity is correlated with relevance. After snippeting, the document context is routed to an LLM inference server.

\subsection{Generation and Attention Output} RAGViz's system requires a node with access to GPUs, represented in Figure \ref{fig: architecture} in green, to run LLM inference tasks. As a first prototype, RAGViz's system uses two model libraries. vLLM \citep{10.1145/3600006.3613165} is a library for fast LLM inference. vLLM is used in RAGViz to efficiently generate text from a prompt created by combining the document context and the query.
Since vLLM does not support attention output, the system then uses the HuggingFace model library \citep{wolf-etal-2020-transformers} to pass both input tokens (document context and query) and output tokens (text generated by vLLM) through the language model and retrieve attention scores. These scores are averaged across all heads and layers for the document window to calculate cumulative document-level attention scores.

\subsection{Frontend User Interface} The frontend user interface is adapted from Search with Lepton \citep{searchbylepton} and uses the Next.JS framework \citep{nextjs}. The frontend is built and exported as static files, which are hosted on an Apache web server \citep{612229}. The frontend utilizes a form to collect query information and other parameters to route to the main backend node.

Once the attention scores are received from the backend, they are stored in React states for use in the attention visualization. As users drag to select output tokens, the system stores a React state that lists the selected token indices. For every output token, the frontend sums the corresponding document token attentions and highlights the relevant, high-attention tokens in the document. The frontend also provides buttons for toggling document inclusion and routes new queries with updated sets of documents to a rewrite endpoint.

\section{Experiment}
This section introduces the chosen configurations of RAGViz's system demonstration and presents efficiency evaluations.
\begin{table*}
\centering
\small
\begin{tabular}{lll}
\hline
\textbf{Function} & \textbf{Median latency (s)} & \textbf{95th percentile latency (s)} \\
\hline
Embedding model and tokenizer & 0.1415 & 0.1609 \\
Single approximate nearest neighbor search call & 0.0654 & 0.0713 \\
Total ANN search and rerank time & 0.0709 & 0.0769 \\
Fetching documents from embedding indices & 0.6092 & 1.0476 \\
Naive first snippeting & 9.1099e-4 & 1.1354e-3 \\
Model generation from vLLM & 1.4571 & 2.3269 \\
Forward pass for attention outputs & 1.1862 & 1.7459 \\
\hline
\textbf{Total query time} & \textbf{5.3923} & \textbf{7.1314} \\
\hline
\end{tabular}
\caption{\label{tab:latency_stats}Latency benchmarking. Latency was measured by executing 50 small general knowledge queries on a RAGViz system that uses the Pile-CC dataset as the data store. The queries have roughly 11 tokens on average.}
\end{table*}
\subsection{Datasets and Settings} RAGViz's demonstration is configured with the following systems:

\textit{Dataset:} RAGViz has been tested with ClueWeb22 \citep{10.1145/3477495.3536321} and The Pile \citep{gao2020pile800gbdatasetdiverse}. ClueWeb22 is a 10-billion-document dataset collected from information-rich webpages. RAGViz uses the 80 million English documents in Category B, which includes the most frequently visited webpages. The Pile is a dataset primarily used for language model training. RAGViz uses the Pile CC training split, which includes filtered HTML pages from the Common Crawl \citep{commoncrawl}. The Pile is used for the demonstration of RAGViz because of its open-source flexibility.

\textit{Embedding model:} We experimented with Anchor-DR \citep{10.1145/3539618.3592080}, an embedding model trained on a contrastive learning task that matches anchor text (text referencing information from linked pages) to those linked pages.

\textit{ANNS system:} RAGViz uses DiskANN \citep{NEURIPS2019_09853c7f}, an efficient graph-based memory-SSD (Solid State Drive) hybrid indexing ANNS system that maintains state-of-the-art performance in terms of latency and recall. DiskANN allows RAGViz's worker nodes to utilize SSDs to reduce memory consumption when serving the index. 

\textit{Language model:} RAGViz uses Llama-2-7b \citep{touvron2023llama2openfoundation}, an open-source language model developed by Meta. Llama-2-7b is lightweight and is supported by both vLLM and HuggingFace. The output token limit is set to 100 tokens for faster performance.

The system demonstration was hosted and evaluated with the hardware listed in Table \ref{tab:resource_stats}.

\subsection{Efficiency Evaluation} We benchmarked the overall efficiency of RAGViz, comparing the two snippeting techniques it offers. Table \ref{tab:latency_stats} shows that the system provides reasonable query latency when using the naive first snippeting method, with most of the latency stemming from LLM generation and the forward pass.

The sliding window technique offers a slight improvement in context relevance, as measured by the inner product. However, it leads to a significant increase in latency, as shown in Table \ref{tab:metric_stats}. The minor relevance improvement makes it difficult to justify the substantial tradeoff in latency.

 \begin{table}[t]
\small
\begin{tabular}{lll}
\hline
\textbf{Metric} & \textbf{Similarity} & \textbf{Latency (s)} \\
\hline
Naive first & 0.97463 & \textbf{9.1099e-4} \\
Sliding window & \textbf{0.97498} & 8.3699 \\
\hline
\end{tabular}
\caption{\label{tab:metric_stats}Comparison between snippeting methods. Average inner product similarity was measured between normalized query and document snippet vectors from executing 50 small general knowledge queries. Latency is measured by the median latency of these queries.}
\end{table}

\begin{table}[t]
\small
\begin{tabular}{llc}
\hline
\textbf{Node} & \textbf{Num CPU Cores} & \textbf{CPU Memory} \\
\hline
Main & 1 & 40 GB \\
Worker & 12 & 85 GB \\
Web* & 24 & 384 GB \\
GPU & 1 & 40 GB \\
\hline
\end{tabular}

\vspace{10pt}

\begin{tabular}{ll}
\hline
\textbf{Node} & \textbf{CPU Type} \\
\hline
Main & Intel® Xeon® E5-2640 v3\\
Worker & Intel® Xeon® E5-2630 v3\\
Web* & 2nd Gen Intel® Xeon® Scalable \\
GPU & 1 Intel® Xeon® E5-2620 v4\\
\hline
\end{tabular}

\vspace{10pt}

\begin{tabular}{lll}
\hline
\textbf{Node} & \textbf{Num GPUs} & \textbf{CUDA Memory} \\
\hline
GPU & 1 & 48 GB \\
\hline
\end{tabular}
\vspace{10pt}

\begin{tabular}{ll}
\hline
\textbf{Node} & \textbf{GPU Type} \\
\hline
GPU & Nvidia RTX A6000\\
\hline
\end{tabular}
\caption{\label{tab:resource_stats}Resources used in our experiments. *Web node is shared by multiple systems outside of RAGViz.}
\end{table}

\section{Conclusion}

RAGViz is a powerful diagnostic tool for analyzing and improving RAG pipelines by providing detailed visualizations of attention mechanisms at various levels. Its attention-driven insights help users better understand the relationship between retrieved documents and language model outputs, making it invaluable for identifying hallucinations and enhancing retrieval efficacy.

As an open-source tool under the MIT license, RAGViz is available for research and development. We plan to support custom models in the future, allowing users to evaluate their own language models within the RAG pipeline. Additionally, we aim to improve usability by containerizing services for more efficient deployment and resource management. We will also unify the LLM inference process to use one inference library, leading to further improvements in speed and resource utilization.

\section{Limitations}
While RAGViz provides valuable visualizations of attention scores between generated and retrieved tokens, it assumes that higher attention scores indicate greater relevance and influence during generation. Further research is needed to evaluate the relationship between attention scores and model interpretability to fully determine RAGViz's effectiveness in improving RAG system explainability.

Currently, RAGViz supports only a single language model for generation tasks, limiting its ability to offer comparative insights across models. Adding support for multiple models could offer a more controlled framework for comparative analysis, enhancing the tool's diagnostic capabilities.

\section*{Acknowledgements}
We would like to thank Jamie Callan and Daniel Vosler for helping with the development and hosting of RAGViz. 
\bibliography{anthology,custom}

\appendix
\end{document}